\pdfoutput=1

\documentclass[11pt]{article}

\usepackage{acl}
\usepackage{times}
\usepackage{latexsym}
\usepackage[T1]{fontenc}
\usepackage[utf8]{inputenc}
\usepackage{microtype}
\usepackage{inconsolata}
\usepackage{graphicx}
\usepackage{pdflscape}
\usepackage{adjustbox}
\usepackage{scalerel}
\usepackage{csquotes}

\title{Conflicts of Interest in Published NLP Research 2000-2024}

\author{
 \textbf{Maarten Bosten\textsuperscript{1}}, 
 \textbf{ Bennett Kleinberg\textsuperscript{1,2}}
\\
\\
 \textsuperscript{1}Tilburg University \\
 \textsuperscript{2}University College London
 \\
 \small{
    {\fontfamily{qcr}\selectfont
        \{m.n.m.l.bosten, bennett.kleinberg\}@tilburguniversity.edu
    }
 }
}

\begin{document}
\maketitle

\begin{abstract}
Natural Language Processing research is increasingly reliant on large scale data and computational power. Many achievements in the past decade resulted from collaborations with the tech industry. But an increasing entanglement of academic research and industry interests leads to conflicts of interest. We assessed published NLP research from 2000-2024 and labeled author affiliations as academic or industry-affiliated to measure conflicts of interest. Overall 27.65\% of the papers contained at least one industry-affiliated author. That figure increased substantially with more than 1 in 3 papers having a conflict of interest in 2024. We identify top-tier  venues (ACL, EMNLP) as main drivers for that effect. The paper closes with a discussion and a simple, concrete suggestion for the future.
\end{abstract}

\section{Introduction}

Recent landmark advancements in Natural Language Processing (NLP) research share the need for an increasing demand for computation and data. Often these demands are hard to meet for individual researchers or university labs, which opens the door for industry collaborations. The involvement of the private sector has led to key publications in its own right \citep[e.g., BERT,][]{devlinBERTPretrainingDeep2019} and has enabled other major achievements through university-industry collaborations \citep[e.g., RoBERTa,][]{liu2019robertarobustlyoptimizedbert}.

However, lessons from other disciplines call for more attention to the role of industry involvement and potential conflicts of interest. In medical research, an entanglement of practitioners and pharmaceutical companies reportedly often materializes in biased research outcomes, impacted on the research agenda, and increased willingness to adopt drugs from these companies \citep{deangelisConflictInterestPublic2000, chrenPhysiciansBehaviorTheir1994,lundhIndustrySponsorshipResearch2017, mccomasResearcherViewsFunding2012}. Current NLP research shares similarities with medical research that make it particularly susceptible to conflicts of interest: both areas are increasingly resource-dependent and applications of academic research in these fields yield substantial benefits to private companies. Gaining insight into the role of industry involvement in NLP research and potential conflicts of interest is as timely as it is lacking. This paper maps out authors' conflicts of interest in NLP research since 2000.

\subsection{Defining conflicts of interest}

There is a conflict of interest if a person P "\textit{is in a relationship with another, requiring P to exercise judgment in the other's behalf [and] has a (special) interest tending to interfere with the proper exercise of judgment in that relationship [where] an interest is any influence, loyalty, concern, emotion, or other feature of a situation tending to make P's judgment (in that situation) less reliable than it would normally be} \citep[][p. 9]{davisConflictInterestProfessions2001}. Importantly, according to this definition having a conflict of interest \textit{does not} imply that person P acts upon the conflicting interest \citep{brodyClarifyingConflictInterest2011}. Whenever an academic researcher (also) holds an affiliation to a private company, there are \textit{by definition} conflicts of interest for that academic researcher. This may be outright undesirable or merely worth mentioning but it matters.

\subsection{Industry involvement in academic research}

A recent study examined institutional affiliations in 22 million scientific articles \citep{hottenrottRiseMultipleInstitutional2021}. For authors with multiple affiliations, the study found that authors affiliated with both a university and company were uncommon across research disciplines (between < 3\% - 4.6\%). Among authors with a single affiliation, 3.7\% held a company affiliation \citep[][supplementary Table A6]{hottenrottRiseMultipleInstitutional2021}.\footnote{These proportions should be interpreted with some caution because more than 20\% of their observations were considered unknown affiliations.}

There are legitimate motivations for academic researchers and industry partners to collaborate. For example, academic researchers might gain access to proprietary and large datasets, expensive material, and have an increased impact on society \citep{perkmannAcademicEngagementCommercialisation2013}. For companies, university-collaborations offer an opportunity to access public funding \citep{perkmannAcademicEngagementCommercialisation2013,bodasfreitasMotivationsInstitutionsOrganization2017}, bolster public trust, increase the perceived value of the company \citep{rothensteinCompanyStockPrices2011,colomboSignalingSciencebasedIPOs2019}, and provide access to talents as potential future employees \citep{ankrahUniversitiesIndustryCollaboration2015,bodasfreitasMotivationsInstitutionsOrganization2017}. 

\subsection{Why should we care?}

Conflicts of interest can skew research findings and increase publication bias more broadly. Ample evidence from medical research suggests that studies that are financially supported by pharmaceutical companies tend to result in outcomes favoring these companies \citep{lundhIndustrySponsorshipResearch2017}. Furthermore, a cross-sectional study of randomized controlled trials found the financial ties of principle investigators to a company to corresponded to a higher likelihood of reporting positive study outcomes \citep{ahnFinancialTiesPrincipal2017}. A systematic analysis of nutritional science literature indicated that studies fully funded by industry were 7x more likely to report favorable versus unfavorable conclusions in comparison to studies who had no industry funding \citep{lesserRelationshipFundingSource2007}.\footnote{Non-financial conflicts of interest could also affect research output. Such conflicts of interest refer to personal relationships or connections with companies \citep[][p. 194]{boutronConsideringBiasConflicts2019}.} 

Another way in which conflicts of interest could undermine scientific output is by presenting results in a self-serving manner. For example, presenting findings with more certainty than warranted could enhance the perceived credibility of a company, while results presented with more uncertainty could protect a company’s interests or competitive advantage \citep{tallapragadaAwareIgnorantExploring2017}.

Others have added that the risks of industry involvement depend on the reliance on resources \citep{ankrahUniversitiesIndustryCollaboration2015}. Access to resources was reported as a strong motivator for universities to collaborate with industry. An increasing reliance on data and compute make NLP research dependent on industry collaborations and this dependency makes universities vulnerable. Shedding light on the presence of conflict of interest in NLP is critical to safeguard the quality of its scientific output. 

\subsection{Aims of this work}

This paper has two central contributions: (1) By resorting to the ACL anthology, we build and provide a dataset for research into conflicts of interest that is near complete for the discipline, which is practically infeasible in other areas with more scattered publication pathways. (2) We conduct analyses about the temporal development of conflicts of interest and the role of publication venues.

\section{Data}

We sought to obtain a dataset of all papers published in the ACL Anthology since 2000\footnote{We targeted papers in official ACL venues: AACL (Asia-Pacific chapter of ACL), ACL, ANLP (Applied NLP), CL (Computational Linguistics journal), CoNLL (Natural Language Learning), EACL (European chapter), EMNLP, IWSLT (Int. Conference on Spoken Language Translation), NAACL (North-American chapter), SemEval, *SEM (Lexical and Computational
Semantics), TACL (Transactions of ACL journal), WMT (Conference on Machine Translation)}. We used the Beautiful Soup \citep{richardsonBeautifulsoup4ScreenscrapingLibrary}, Requests and Asyncio libraries in Python 3.10 to download all published papers since 2000 and converted the pdf papers to XML with GROBID \citep{GROBID}. From the XML tree, we extracted key variables (e.g., author affiliation, year, venue) using ElementTree XML \citep{XmlEtreeElementTree} and structured it in a data frame using pandas \citep{thepandasdevelopmentteamPandasdevPandasPandas2023}. All statistical analyses were done in R \citep{rcoreteamLanguageEnvironmentStatistical2023}.\footnote{The final dataset, an extended version with abstracts and full papers, the code to recreate the dataset and the analysis code are available at: \url{https://osf.io/gfqpr/?view_only=2a45d1a34e8e4a96a26c8cf5ed324ca8}}

That procedure resulted in $n=66924$ papers, of which 502 we not properly converted to XML. An additional 1,110 papers were excluded due to error handling or missing data \footnote{Observations were excluded at the author level since this is our unit of interest.}. We then excluded 3,687 papers due to authors without an affiliation or incorrectly parsed affiliations (e.g., due to encoding issues). After aggregation to the paper-level, we excluded collections or entire proceedings files of all papers published at a workshop/venue ($n=2275$ papers). Lastly, we excluded papers with more than 100+ references or authors ($n=663$) to reduce incorrectly formatted papers and leftover conference proceedings. The final dataset consisted of $n=58687$ papers.

\subsection{Affiliation labeling}
\noindent We specified four affiliation patterns used for labeling and engaged in an iterative process of labeling and inspecting affiliations. Affiliations were labeled as \textit{industry}, \textit{non-industry}, \textit{other}, and \textit{unsure}. Industry affiliations were those that  could reasonably be assumed to be for-profit companies. The corresponding pattern consisted of common corporate abbreviations (e.g., ltd, GmbH, Corporation, corp.) and brand names (e.g., Meta, OpenAI, Uber). Affiliations were labeled as non-industry if they referred to a university, research institute, or non-profit company. The corresponding pattern consisted of abbreviations associated with scientific or non-profit institutions (e.g., University, FBK, US Naval Academy). Affiliations that could not be assigned to either industry or non-industry were labeled as \textit{other} and not considered for further analysis. Unsure affiliations (i.e., STAR, LIMSI) were considered non-industry. Of all affiliations, 99.62\% were assigned to one of the four patterns.

\section{Findings}

Overall, 27.65\% ($n=16226)$ of all papers had at least one author with an industry affiliation. Conversely, 72.35\% ($n=42461$) of the papers stem from authors with solely academic affiliations. Among the industry-affiliated papers, 1,574 papers were published exclusively from industry-affiliated authors (2.68\% of all papers).

There was substantial variation in the proportion of industry affiliations over the publications venues considered.\footnote{ANLP was discontinued after 2000. The results were not affected by in- or exclusion, so ANLP was retained in the dataset.} Table \ref{tab:my-table} indicates that industry affiliation in the dataset is driven by a few venues who attract a disproportionately large number of industry papers. A Chi-square test suggested that there was a significant association between a paper being academic or industry-affiliated and the venue, $\chi^2(13)=1232.5, p<.001$. To identify the drivers of that association, we looked at the standardized residuals (z-scores).\footnote{Absolute z-scores > 2.33 correspond to p < .01, two-tailed.} Positive z-scores imply that the venue has a larger proportion of industry papers than would be expected if there were no association between affiliation and venue. 

By far the largest drivers of the relationship were EMNLP and ACL followed by NAACL and TACL. In contrast, workshops, SemEval, Starsem, EACL and the ACL journal Computational Linguistics (CL) contained significantly fewer industry papers than would have been expected. It is noteworthy that the over-representation of industry-affiliated papers follows regional conference patterns: the North-American chapter (NAACL) has substantially higher proportions of industry papers than the European (EALP) or Asia-Pacific one (AACL). Moreover, among journals, CL attracts fewer industry papers than TACL, arguably due to different foci, with CL being more embedded in \textit{traditional} computational linguistics. The disparity between workshops (including SemEval and Starsem) and the ACL main conferences is considerable and might reflect competing ambitions (i.e., a narrow focus and depth for workshops versus broad technical advancements at large venues), which is exemplified by EMNLP being by far the most attractive outlet for industry papers with more than 1/3 industry papers.

\begin{table}
\centering
\begin{tabular}{lccr}
\hline
        & \textbf{Academic} & \textbf{Industry} & \textbf{z-score} \\
\hline
AACL    & 0.70     & 0.30     & 1.39   \\
ACL     & 0.68     & 0.32     & 12.63   \\
ANLP    & 0.73     & 0.27     & -0.08   \\
CL      & 0.77     & 0.23     & -2.94   \\
CoNLL   & 0.74     & 0.26     & -0.75   \\
EACL    & 0.76     & 0.24     & -4.99   \\
EMNLP   & 0.65     & 0.35     & 21.16   \\
IWSLT   & 0.74     & 0.26     & -0.63   \\
NAACL   & 0.70     & 0.30     & 4.59    \\
SemEval & 0.79     & 0.21     & -7.51   \\
StarSem & 0.81     & 0.19     & -4.21   \\
TACL    & 0.68     & 0.32     & 2.49    \\
WMT     & 0.73     & 0.27     & -0.17    \\
WS      & 0.82     & 0.18     & -29.14 \\
\hline
\end{tabular}
\caption{Proportion of affiliation type by venue with z-scores (standardized residuals). Positive z-scores imply a higher proportion of industry papers than expected if there were no association, and vice versa for negative z-scores.}
\label{tab:my-table}
\end{table}

\begin{figure*}
    \centering
    \includegraphics[width=1\linewidth]{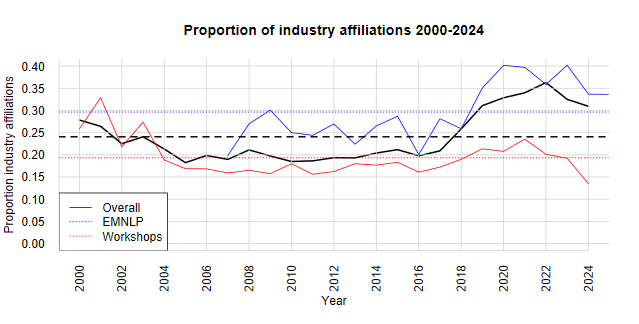}
    \caption{Industry affiliations 2000-2024. Industry papers are those that contain at least one industry-affiliated authors. Dashed (dotted) lines indicate the mean across the time span for the respective group.}
    \label{fig:time-trend}
\end{figure*}

When assessed over time (Fig. \ref{fig:time-trend}), we find evidence for an acceleration of industry collaborations. Across all venues, there was an initial decrease and between 2002 and 2018, the proportion of industry-affiliated papers remained below 0.25. From 2018 onward, the proportion increased and remained over 0.25 with a peak in 2022 (0.36). For reference, Fig. \ref{fig:time-trend} shows the two primary drivers for either direction\footnote{Non-missing time series data for EMNLP are available from 2007 onward.}. From 2016 onward, both venue types diverge, mainly driven by stark increases in industry papers at EMNLP.

\section{Discussion}

Conflicts of interest are important to understand since NLP research and private sector interests are intertwined at an accelerating rate. Our findings indicate that 1 in 3 peer-reviewed papers in NLP research contain a conflict of interest. Top venues such as EMNLP and ACL attract a disproportionate amount of these works, which may reflect the quality and scope of the work presented in industry-affiliated papers or be indicative of publication preferences, or both.  

Our work does not take a normative stance regarding conflicts of interest; the short- and long-term research implications are yet to be determined. As such, we do not suggest that having a conflict of interest is problematic, \textit{per se}. For many researchers, substantial remunerations and access to resources are convincing reasons for industry collaboration. 

However, both the proportion of industry papers and the findings about the consequences from other disciplines warrant caution. When scientific rigor and private interests are at odds, this it could result in selective reporting \citep[i.e., under-reporting disappointing or null findings,][]{ahnFinancialTiesPrincipal2017, lundhIndustrySponsorshipResearch2017}, drifting research agendas \citep[e.g., focused on commercial potential,][]{mccomasResearcherViewsFunding2012} and abandonment of critical research lines \citep{nytimesGoogleResearcher, 10.1145/3442188.3445922}.

\noindent The exact nature of conflicts of interest on scientific quality might be more nuanced than merely leading to publication bias and favorable outcomes \citep{bodasfreitasMotivationsInstitutionsOrganization2017, ankrahUniversitiesIndustryCollaboration2015}. Some work suggests an inverted U-shaped relationship between industry funding and research output metrics \citep{muscioComplexRelationshipAcademic2017}: industry funding correlated positively to research output up until an inflection point, after which the benefits vanished. That effect was discipline-specific, with the medical and health sciences only benefiting from industry-funding when funding was substantial.

\subsection{Future work and limitations}
\noindent Out dataset offers the means necessary to explore the role of conflicts of interest at the textual level (e.g., whether papers with industry-affiliated authors cover different topics or present findings differently \citep{boutronMisrepresentationDistortionResearch2018}). Another line of inquiry could assess, at the author-level, whether becoming industry affiliated impacts publication practices in a quasi-experimental design (e.g., comparing research output and practices before and after becoming industry-affiliated). To better understand the impact of conflicts of interest, future work could compare disciplines and formalize variables that measure scientific impact or progress. It is important to emphasize that our work resorted to a literal and strict operationalizations of conflicts of interest and that other operationalizations are also plausible.

\subsection{A (very) simple solution}

With the already high rate of conflicts of interest in NLP research and no sign of that trend disappearing, it is surprising that the solution is rather simple. While common in most scientific disciplines, the ACL has not established a mandatory conflicts of interest statement for authors in its code of ethics\footnote{\url{https://www.aclweb.org/adminwiki/index.php/ACL_Conference_Conflict-of-interest_policy}}. Ironically, the same authors need to declare their conflicts of interests in academic journals but not in ACL venues. A simple statement or checkbox during submission that mentions conflicts of interest and the nature thereof, would make these transparent and go along way towards mitigating adverse effects of the intensified entanglement of scientific and private interests.

\bibliography{coiacl}

\appendix

\section{Venue labeling}
For each paper, we extracted the conference venue through regular expression queries on the papers' URLs. These were papers where their \textit{Tag}, the unique identifier retrieved from the URL, corresponded to a single venue (e.g., D18-1402, all tags starting with the letter D corresponded to the EMNLP venue). The tags varied over the years, but most of them followed a similar pattern.

For venues with the same starting letter in their identifier, we manually checked which range of papers belonged to which venue. The remaining papers had tags that specifically mentioned a venue or workshop (e.g., 2021-nlp4convai-1-9). To identify these papers, we wrote a small script that retrieved the names of all workshops for each venue from the ACL webpage. These were then stored and manually searched to connect the papers to the correct venue.

\section{Supplementary Tables}

\begin{landscape}
  \begin{table}
    \caption{Number of papers across venues and years.}
    \label{table:papersvenueyear}
\begin{adjustbox}{width=1.38\textwidth}
    \begin{tabular}{lrrrrrrrrrrrrrrrrrrrrrrrrrr}
  \hline
CID & 2024 & 2023 & 2022 & 2021 & 2020 & 2019 & 2018 & 2017 & 2016 & 2015 & 2014 & 2013 & 2012 & 2011 & 2010 & 2009 & 2008 & 2007 & 2006 & 2005 & 2004 & 2003 & 2002 & 2001 & 2000 & Total \\ 
  \hline
AACL & 53 & 164 & 271 & - & 188 & - & - & - & - & - & - & - & - & - & - & - & - & - & - & - & - & - & - & - & - & 676 \\ 
  ACL & 2284 & 2494 & 1473 & 1461 & 1073 & 726 & 421 & 342 & 363 & 344 & 317 & 385 & 218 & 327 & 256 & 233 & 208 & 191 & 290 & 127 & 125 & 98 & 51 & 64 & 40 & 13911 \\ 
  ANLP & - & - & - & - & - & - & - & - & - & - & - & - & - & - & - & - & - & - & - & - & - & - & - & - & 88 & 88 \\ 
  CL & 18 & 22 & 28 & 20 & 13 & 17 & 29 & 17 & 27 & 30 & 33 & 34 & 36 & 42 & 38 & 30 & 33 & 35 & 38 & 29 & 22 & 26 & 25 & 40 & 38 & 720 \\ 
  CoNLL & 35 & 67 & 24 & 52 & 59 & 110 & 88 & 79 & 51 & 51 & - & - & - & - & - & - & - & - & - & - & - & - & - & - & - & 616 \\ 
  EACL & 704 & 679 & 20 & 685 & 40 & - & - & 275 & - & - & 160 & - & 113 & - & - & 124 & - & - & 85 & - & - & 106 & - & - & - & 2991 \\ 
  EMNLP & 2580 & 2468 & 1813 & 1617 & 1601 & 1033 & 553 & 340 & 256 & 299 & 223 & 200 & 134 & 141 & 123 & 156 & 113 & 126 & - & 141 & - & - & - & - & - & 13917 \\ 
  IWSLT & 32 & 41 & 29 & 29 & 26 & 31 & 26 & 18 & 25 & - & 40 & 41 & 41 & 36 & 20 & 24 & 24 & 22 & 26 & 24 & 19 & - & - & - & - & 574 \\ 
  NAACL & 1171 & - & 1000 & 860 & 44 & 471 & 389 & - & 212 & 232 & - & 164 & 113 & - & 169 & 166 & - & 144 & 132 & - & 92 & 92 & - & 16 & - & 5467 \\ 
  SemEval & 255 & 275 & 209 & 176 & 269 & 216 & 185 & 174 & 201 & 152 & 141 & 110 & - & - & 97 & - & - & 105 & - & - & - & - & - & 33 & - & 2598 \\ 
  StarSem & 34 & 44 & 28 & 28 & 19 & 30 & 30 & 29 & 27 & 35 & 21 & 44 & 104 & - & - & - & - & - & - & - & - & - & - & - & - & 473 \\ 
  TACL & 69 & 87 & 77 & 86 & 51 & 42 & 47 & 34 & 39 & 40 & 42 & 33 & - & - & - & - & - & - & - & - & - & - & - & - & - & 647 \\ 
  WMT & 112 & 83 & 111 & 102 & 124 & 102 & 101 & 73 & 98 & 60 & 59 & 60 & 54 & 63 & 62 & 39 & 33 & 33 & 23 & - & - & - & - & - & - & 1392 \\ 
  WS & 475 & 354 & 189 & 17 & 592 & 1265 & 1027 & 1161 & 1013 & 945 & 945 & 845 & 758 & 749 & 690 & 692 & 333 & 346 & 499 & 255 & 488 & 318 & 275 & 146 & 240 & 14617 \\ 
  \hline
  Total & 7822 & 6778 & 5272 & 5133 & 4099 & 4043 & 2896 & 2542 & 2312 & 2188 & 1981 & 1916 & 1571 & 1358 & 1455 & 1464 & 744 & 1002 & 1093 & 576 & 746 & 640 & 351 & 299 & 406 & 58687 \\ 
   \hline
    \end{tabular}
  \end{adjustbox}
\end{table}
\end{landscape}

\begin{landscape}
  \begin{table}
    \caption{Average number of authors across venues and years.}
    \label{table:authorsvenueyear}
\begin{adjustbox}{width=1.38\textwidth}
  \begin{tabular}{lrrrrrrrrrrrrrrrrrrrrrrrrrr}
  \hline
CID & 2024 & 2023 & 2022 & 2021 & 2020 & 2019 & 2018 & 2017 & 2016 & 2015 & 2014 & 2013 & 2012 & 2011 & 2010 & 2009 & 2008 & 2007 & 2006 & 2005 & 2004 & 2003 & 2002 & 2001 & 2000 & Average \\ 
  \hline
AACL & 3.66 & 4.04 & 3.42 & - & 3.39 & - & - & - & - & - & - & - & - & - & - & - & - & - & - & - & - & - & - & - & - & 3.63 \\ 
  ACL & 4.94 & 4.48 & 3.96 & 3.97 & 3.61 & 3.70 & 3.35 & 3.23 & 3.13 & 3.17 & 2.88 & 2.87 & 2.79 & 2.55 & 2.64 & 2.70 & 2.50 & 2.48 & 2.54 & 2.22 & 2.26 & 3.13 & 2.02 & 2.25 & 2.15 & 3.02 \\ 
  ANLP & - & - & - & - & - & - & - & - & - & - & - & - & - & - & - & - & - & - & - & - & - & - & - & - & 1.99 & 1.99 \\ 
  CL & 2.89 & 3.55 & 3.11 & 2.75 & 2.92 & 3.29 & 3.03 & 4.59 & 3.85 & 4.27 & 4.30 & 4.50 & 4.44 & 3.38 & 3.00 & 3.77 & 3.33 & 2.49 & 2.11 & 2.79 & 2.23 & 2.62 & 2.16 & 3.17 & 3.84 & 3.30 \\ 
  CoNLL & 3.60 & 3.87 & 3.25 & 3.67 & 2.92 & 3.65 & 3.02 & 3.49 & 3.41 & 2.96 & - & - & - & - & - & - & - & - & - & - & - & - & - & - & - & 3.38 \\ 
  EACL & 3.94 & 4.01 & 2.70 & 3.26 & 3.17 & - & - & 3.06 & - & - & 2.64 & - & 2.63 & - & - & 2.32 & - & - & 2.22 & - & - & 2.39 & - & - & - & 2.94 \\ 
  EMNLP & 4.91 & 4.54 & 4.35 & 3.82 & 3.80 & 3.69 & 3.48 & 3.17 & 3.35 & 3.05 & 2.88 & 2.96 & 3.04 & 2.93 & 2.81 & 2.65 & 2.47 & 2.55 & - & 2.67 & - & - & - & - & - & 3.32 \\ 
  IWSLT & 8.09 & 5.78 & 6.07 & 4.48 & 4.69 & 3.84 & 3.96 & 3.56 & 4.20 & - & 4.72 & 4.83 & 4.59 & 4.39 & 3.40 & 2.92 & 3.71 & 3.14 & 2.85 & 3.42 & 3.47 & - & - & - & - & 4.31 \\ 
  NAACL & 4.98 & - & 4.02 & 3.82 & 3.86 & 3.41 & 3.23 & - & 3.37 & 2.85 & - & 2.66 & 2.47 & - & 2.53 & 2.54 & - & 2.53 & 2.50 & - & 3.12 & 2.45 & - & 1.50 & - & 3.05 \\ 
  SemEval & 3.90 & 3.72 & 3.55 & 3.64 & 3.11 & 3.23 & 3.18 & 3.28 & 3.20 & 3.31 & 3.38 & 2.96 & - & - & 2.81 & - & - & 2.53 & - & - & - & - & - & 3.24 & - & 3.27 \\ 
  StarSem & 3.00 & 3.23 & 3.07 & 3.64 & 2.84 & 3.37 & 2.60 & 3.17 & 2.59 & 3.26 & 2.48 & 3.09 & 2.88 & - & - & - & - & - & - & - & - & - & - & - & - & 3.02 \\ 
  TACL & 4.64 & 4.08 & 4.64 & 3.88 & 3.51 & 3.43 & 3.06 & 3.26 & 2.85 & 2.80 & 2.71 & 2.33 & - & - & - & - & - & - & - & - & - & - & - & - & - & 3.43 \\ 
  WMT & 4.35 & 4.71 & 5.25 & 4.76 & 3.92 & 3.73 & 3.64 & 3.82 & 3.47 & 3.98 & 4.08 & 3.67 & 3.15 & 3.65 & 2.97 & 4.10 & 3.42 & 2.55 & 2.70 & - & - & - & - & - & - & 3.79 \\ 
  WS & 3.26 & 3.36 & 3.33 & 2.76 & 3.10 & 3.09 & 3.05 & 2.93 & 2.94 & 2.85 & 2.77 & 2.87 & 2.67 & 2.66 & 2.77 & 2.56 & 2.51 & 2.55 & 2.53 & 2.29 & 2.56 & 2.55 & 2.37 & 2.27 & 2.13 & 2.75 \\ 
  \hline
  Average & 4.32 & 4.11 & 3.90 & 3.71 & 3.45 & 3.49 & 3.24 & 3.42 & 3.30 & 3.25 & 3.29 & 3.28 & 3.19 & 3.26 & 2.87 & 2.95 & 2.99 & 2.60 & 2.49 & 2.68 & 2.73 & 2.63 & 2.18 & 2.49 & 2.53 & 3.23 \\ 
   \hline
  \end{tabular}
\end{adjustbox}
\end{table}
\end{landscape}

\begin{landscape}
  \begin{table}
    \caption{Average number of references across venues and years.}
    \label{table:referencesvenueyear}
\begin{adjustbox}{width=1.38\textwidth}
  \begin{tabular}{lrrrrrrrrrrrrrrrrrrrrrrrrrr}
  \hline
CID & 2024 & 2023 & 2022 & 2021 & 2020 & 2019 & 2018 & 2017 & 2016 & 2015 & 2014 & 2013 & 2012 & 2011 & 2010 & 2009 & 2008 & 2007 & 2006 & 2005 & 2004 & 2003 & 2002 & 2001 & 2000 & Average \\ 
  \hline
AACL & 30.4 & 39.7 & 35.1 & - & 29.5 & - & - & - & - & - & - & - & - & - & - & - & - & - & - & - & - & - & - & - & - & 33.7 \\ 
  ACL & 43.5 & 41.6 & 39.4 & 37.9 & 34.6 & 34.7 & 31.8 & 30.9 & 30.5 & 28.6 & 27.5 & 25.5 & 22.6 & 22.0 & 23.0 & 17.2 & 17.2 & 15.9 & 17.9 & 16.0 & 14.7 & 14.7 & 15.9 & 17.3 & 15.4 & 25.5 \\ 
  ANLP & - & - & - & - & - & - & - & - & - & - & - & - & - & - & - & - & - & - & - & - & - & - & - & - & 15.2 & 15.2 \\ 
  CL & 51.9 & 44.8 & 52.2 & 49.0 & 54.0 & 60.2 & 39.3 & 42.1 & 42.6 & 34.4 & 49.6 & 48.1 & 34.7 & 28.6 & 26.1 & 34.1 & 27.9 & 19.2 & 22.2 & 27.1 & 28.6 & 34.7 & 37.7 & 19.1 & 22.3 & 37.2 \\ 
  CoNLL & 44.0 & 39.1 & 44.7 & 39.8 & 38.6 & 37.8 & 31.2 & 29.7 & 27.1 & 26.6 & - & - & - & - & - & - & - & - & - & - & - & - & - & - & - & 35.9 \\ 
  EACL & 36.9 & 39.7 & 33.2 & 33.6 & 29.8 & - & - & 28.2 & - & - & 24.2 & - & 24.3 & - & - & 22.1 & - & - & 15.5 & - & - & 13.3 & - & - & - & 27.4 \\ 
  EMNLP & 44.2 & 42.4 & 40.8 & 38.6 & 36.1 & 30.7 & 32.7 & 32.1 & 32.8 & 28.4 & 29.7 & 28.6 & 30.6 & 31.3 & 25.6 & 24.5 & 22.8 & 23.4 & - & 16.5 & - & - & - & - & - & 31.1 \\ 
  IWSLT & 29.8 & 32.9 & 33.1 & 31.2 & 25.2 & 27.5 & 25.7 & 25.3 & 23.6 & - & 24.4 & 25.7 & 18.9 & 19.4 & 18.9 & 14.2 & 17.1 & 18.0 & 17.5 & 15.3 & 14.2 & - & - & - & - & 22.9 \\ 
  NAACL & 40.9 & - & 38.8 & 34.6 & 19.6 & 33.2 & 30.9 & - & 28.8 & 27.1 & - & 24.4 & 22.7 & - & 18.1 & 16.3 & - & 14.1 & 14.0 & - & 14.3 & 11.4 & - & 17.6 & - & 24.0 \\ 
  SemEval & 21.6 & 20.8 & 22.3 & 22.2 & 21.4 & 18.7 & 18.2 & 17.3 & 18.5 & 16.9 & 16.2 & 16.6 & - & - & 10.5 & - & - & 9.2 & - & - & - & - & - & 4.6 & - & 17.0 \\ 
  StarSem & 41.2 & 40.9 & 39.9 & 35.4 & 36.5 & 28.7 & 32.2 & 30.7 & 28.5 & 26.1 & 27.6 & 22.9 & 17.9 & - & - & - & - & - & - & - & - & - & - & - & - & 31.4 \\ 
  TACL & 59.4 & 54.2 & 59.4 & 53.9 & 51.4 & 50.0 & 50.1 & 46.5 & 43.4 & 42.5 & 41.4 & 38.0 & - & - & - & - & - & - & - & - & - & - & - & - & - & 49.2 \\ 
  WMT & 27.1 & 29.3 & 27.3 & 25.2 & 24.6 & 23.2 & 24.9 & 24.3 & 21.4 & 23.8 & 26.3 & 25.5 & 21.8 & 20.0 & 17.5 & 16.3 & 15.2 & 16.2 & 14.5 & - & - & - & - & - & - & 22.3 \\ 
  WS & 29.3 & 25.4 & 27.5 & 30.0 & 25.3 & 27.2 & 25.2 & 23.6 & 23.0 & 22.1 & 20.2 & 20.6 & 19.0 & 18.1 & 16.6 & 16.2 & 14.9 & 15.8 & 16.2 & 16.6 & 14.4 & 14.5 & 15.0 & 14.0 & 13.6 & 20.2 \\ 
  \hline
  Average & 38.5 & 37.6 & 38.0 & 36.0 & 32.8 & 33.8 & 31.1 & 30.1 & 29.1 & 27.7 & 28.7 & 27.6 & 23.6 & 23.2 & 19.5 & 20.1 & 19.2 & 16.5 & 16.8 & 18.3 & 17.2 & 17.7 & 22.8 & 14.5 & 16.6 & 28.1 \\ 
   \hline
  \end{tabular}
\end{adjustbox}
\end{table}
\end{landscape}

\section{Supplementary Images}

\begin{figure}[h]
    \begin{center}
    \includegraphics[width=15.5cm]{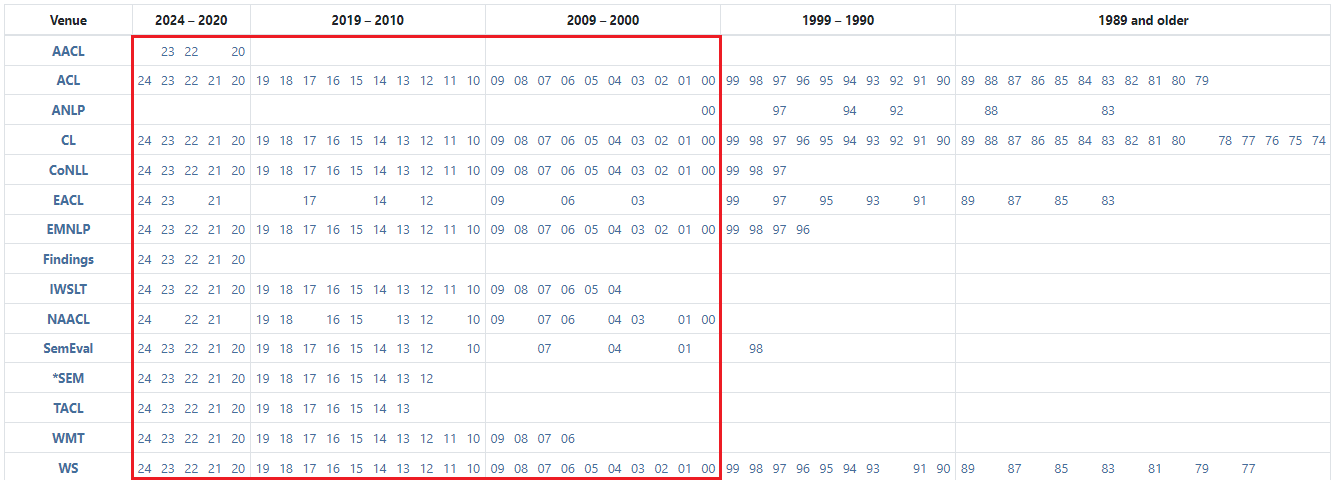}
    \caption{Visual representation of all conferences that are included in the dataset.}
    \label{fig:conferenceselection}
    \end{center}
\end{figure}

\end{document}